\RequirePackage[T1]{fontenc}
\documentclass[10pt,twocolumn,letterpaper]{article}
\usepackage[pagenumbers]{cvpr}
\usepackage{amsmath,amssymb}
\usepackage{microtype}
\definecolor{cvprblue}{rgb}{0.21,0.49,0.74}
\usepackage[pagebackref,breaklinks,colorlinks,allcolors=cvprblue]{hyperref}
\hypersetup{pdfauthor={},pdftitle={Boltz-Flow: Generating High-Quality Robot Action from Boltzmann Machine}}
\def\confName{CVPR}
\def\confYear{2027}
\title{Boltz-Flow: Generating High-Quality Robot Action from Boltzmann Machine}
\author{Anonymous Authors}

\usepackage{xcolor}

\author{
\parbox{\textwidth}{\centering\footnotesize
\mbox{Hengzhi Jiang\textsuperscript{2,$\dagger$}},
\mbox{Haixu Liu\textsuperscript{1,$\dagger$}},
\mbox{Jindou Jia\textsuperscript{3,$\dagger$}},
\mbox{Tinglan Wang\textsuperscript{1,$\dagger$}},
\mbox{Shaobo Chen\textsuperscript{1,$\dagger$}},
\mbox{Yiren Zhao\textsuperscript{4,$\dagger$}},
\mbox{Ziyang Rao\textsuperscript{4}},
\mbox{Laiyuan Cui\textsuperscript{1}},
\mbox{Jiaxin Dong\textsuperscript{4}},
\mbox{Yin Ma\textsuperscript{1}},
\mbox{Hai Wei\textsuperscript{1}},
\mbox{Jiaqi Bai\textsuperscript{3}},
\mbox{Tuo An\textsuperscript{3}},
\mbox{Xinyu Zhou\textsuperscript{3}},
\mbox{Hao Ban\textsuperscript{1}},
\mbox{Hongdong Zhu\textsuperscript{1}},
\mbox{Yanqi Wang\textsuperscript{3}},
\mbox{Jiaqi Dou\textsuperscript{2}},
\mbox{Nuo Qian\textsuperscript{1}},
\mbox{Qi Liu\textsuperscript{5}},
\mbox{Bin Jin\textsuperscript{5}},
\mbox{Chang Wu\textsuperscript{1}},
\mbox{Qingfeng Li\textsuperscript{2}},
\mbox{Hui Yang\textsuperscript{5}},
\mbox{Weiyu Guo\textsuperscript{6,*}},
\mbox{Dicong Qiu\textsuperscript{5,*}},
\mbox{Qi Gao\textsuperscript{1,*}},
\mbox{Jianwei Niu\textsuperscript{2,*}},
\mbox{Kai Wen\textsuperscript{1,*}},
\mbox{Jianfei Yang\textsuperscript{3,*}}\\[0.4em]
{\scriptsize
\textsuperscript{1}Beijing QBoson Quantum Technology Co., Ltd. \quad
\textsuperscript{2}Beihang University \quad
\textsuperscript{3}Nanyang Technological University \quad
\textsuperscript{4}The Hong Kong University of Science and Technology (Guangzhou) \quad
\textsuperscript{5}Quantsparkle.ai \quad
\textsuperscript{6}The Hong Kong University of Science and Technology
}
}
}

\begin{document}
\maketitle

\begingroup
\renewcommand{\thefootnote}{}
\footnotetext{
\begin{tabular}[t]{@{}l@{\hspace{0.4em}}l@{}}
$\dagger$ & These authors contributed equally to this work. \\
$*$ & Corresponding authors.
\end{tabular}
}
\endgroup

\begin{abstract}

Flow-matching policies integrate a velocity field from a fixed, observation-independent \textit{Gaussian} source. However, since the source is unimodal and identical across all tasks, the limited representational capacity of \textit{Gaussian} distributions constrains how far a policy can scale with growing task diversity. Moreover, the source factorizes over time steps and control dimensions, discarding the coordination that manipulation requires. We introduce \textbf{Boltz-Flow}, which replaces \textit{Gaussian} initialization with a more powerful, observation- and instruction-conditioned \textit{Boltzmann} source while leaving the base VLA unchanged. \textbf{Boltz-Flow} parameterizes this source with a conditional \textit{Boltzmann} machine that assigns an energy value to each binary source state. A listwise objective, supervised entirely by offline demonstrations, aligns this energy with the quality of actions produced after frozen flow integration. At inference, the energy defines an Ising instance whose solution is decoded into continuous noise before a single standard integration. We use a physical coherent Ising machine during training. 
Experiments show that Boltz-Flow improves $\pi_{0.5}$ across three benchmarks, outperforming the compared strong policy baselines and competing noise optimization methods on LIBERO and RoboTwin 2.0 and ranking second only to LEAP on LIBERO-Plus.
Additional experiments show gains under noise-adapted fine-tuning, with action-to-action warm starts, and on the SmolVLA backbone.
\end{abstract}

\section{Introduction}

The long-term goal of robot learning is to enable robots to acquire and execute
general-purpose physical skills in open-world environments. With large-scale
robot datasets and vision-language pretraining, vision-language-action (VLA)
models have become a key paradigm connecting scene perception,
language understanding, and robot control~\cite{kim2025openvla}.  VLAs,
such as RT-2~\cite{zitkovich2023rt} and OpenVLA~\cite{kim2025openvla},
discretize continuous actions into tokens and generate them
autoregressively~\cite{pertsch2025fast}. For continuous action
generation~\cite{jia2026ode}, diffusion policy~\cite{chi2023diffusion} casts
action synthesis as iterative denoising. Flow matching~\cite{lipman2023flow}
instead learns a conditional velocity field that transports a
\textit{Gaussian} source distribution to the action distribution. VLAs such as
$\pi_0$~\cite{black2024pi0} and $\pi_{0.5}$~\cite{physicalintelligence2025pi05}
combine such a flow-matching action expert with a pretrained vision-language
backbone and generate generalizable actions.
\begin{figure*}[t]
  \centering
  \includegraphics[width=0.9\textwidth]{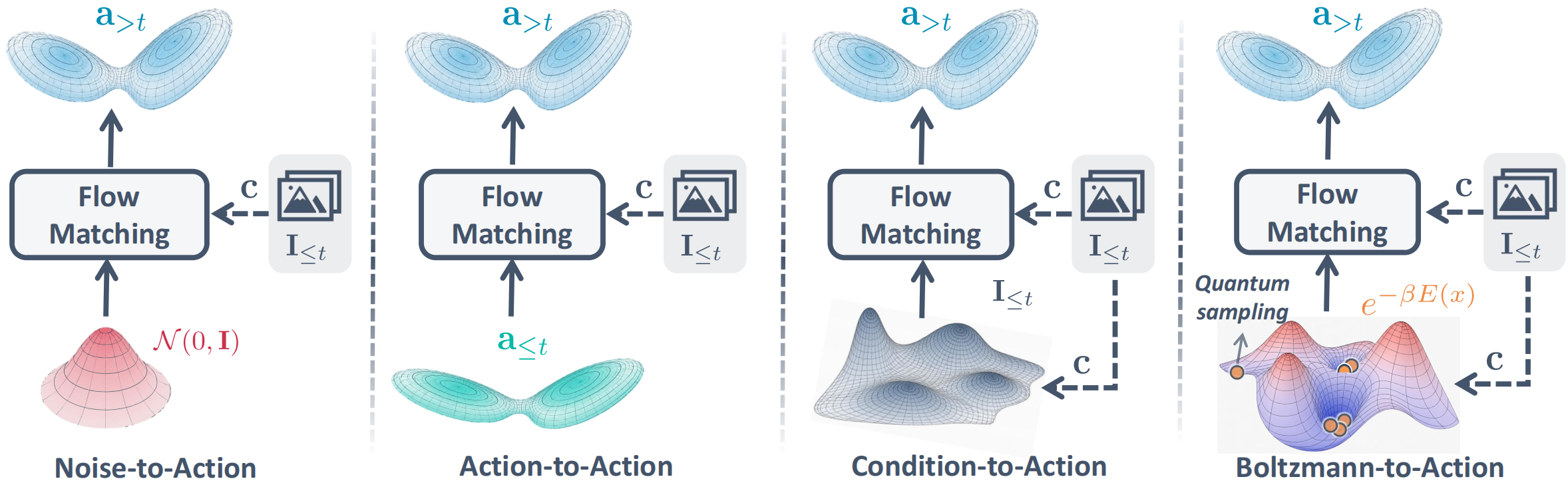}
  \caption{Four choices of source distribution for a flow-matching policy. \textbf{Noise-to-Action}~\cite{lipman2023flow} draws from a fixed isotropic \textit{Gaussian}. \textbf{Action-to-Action}~\cite{jia2026a2a} re-centers the source on past actions. \textbf{Condition-to-Action}~\cite{zhai2025vfp} predicts it from the current observation $\mathbf{I}_{\leq t}$. \textbf{Boltzmann-to-Action} (ours) samples from a learned observation-conditioned \textit{Boltzmann} distribution, whose energy supports dense couplings and multiple low-energy modes, obtained by Ising sampling on a coherent Ising machine.}
  \label{fig:first_fig}
\end{figure*}

However, the \textit{Gaussian} source is specified by design rather than
learned from data, although it fully determines where flow integration
starts~\cite{lipman2023flow}. This limits flow policies in two ways. The first
is capacity. An isotropic \textit{Gaussian} is unimodal and identical for every
task, scene, and instruction, so its probability mass cannot shift toward the
actions that a particular observation calls for. As datasets broaden and
horizons lengthen, the expressiveness of the starting point, rather than the
capacity of the velocity field, likely becomes the limiting factor. The second
is coupling. The source is sampled independently across time steps and control
dimensions, whereas position, orientation, and gripper motion must stay
coordinated across an entire action chunk. Condition-to-action~\cite{zhai2025vfp}
and action-to-action~\cite{jia2026a2a} sources each relax one limits.
However, both narrow the source around a single predicted start and thus trade
representational capacity for conditioning~\cite{jia2026mars}. A better source
should therefore be conditioned on the observation, coupled across variables,
and still expressive enough to hold several action modes.

To this end, we turn to the \textit{Boltzmann} distribution. A \textit{Boltzmann} machine (BM) naturally supports multimodal distributions and captures richer dependencies than a standard \textit{Gaussian}~\cite{florence2022implicit,singh2024revisiting}. Moreover, recent quantum computing devices, such as the coherent Ising machine (CIM)~\cite{wei2026cim}, have renewed interest in applying \textit{Boltzmann} models to deep learning~\cite{bohm2022noise,wang2025quantum}. This raises a fundamental question for robot learning. \emph{Can robot actions be generated from a Boltzmann distribution instead of simple Gaussian noise to achieve stronger manipulation performance?} This work answers it in the affirmative. We propose \textbf{Boltz-Flow}, a \textit{\textbf{Boltz}mann}-based robot \textbf{Flow} policy that generates high-quality actions from a learnable, observation-conditioned, and variable-coupled source distribution (Fig.~\ref{fig:first_fig}). To exploit the action dependencies in demonstrations, Boltz-Flow follows the maximum-entropy principle and builds an instance-specific Gibbs--Boltzmann energy landscape for each observation. This landscape assigns higher probability to low-energy structured sources and makes no unsupported assumptions about the geometry of the remaining action modes.

Concretely, Boltz-Flow instantiates this source with a conditional BM. A multimodal encoder maps the image, instruction, and robot state to a representation that parameterizes a dense quadratic energy over a binary encoding of the initial source. For a fixed observation, this energy is an Ising instance. The CIM searches its low-energy configurations, and its programmable all-to-all connectivity matches the couplings across time steps and action dimensions. We use the CIM for this search during training, and simulated annealing (SA) can replace it at inference to reduce latency. The returned state is decoded into continuous noise, from which the frozen $\pi_{0.5}$ runs its original ten-step integration. The action is thus still produced entirely by the pretrained velocity field. Training uses offline demonstrations alone. We rank candidate sources by how far the resulting frozen-policy actions deviate from the expert action, and a listwise objective aligns the energy landscape with this ranking. At test time, low energy alone identifies a high-quality source.


We evaluate Boltz-Flow on LIBERO~\cite{liu_zhu_NeurIPS2023}, LIBERO-Plus~\cite{Fei_2026_CVPR}, and RoboTwin 2.0. Across the 40 LIBERO tasks, CIM-trained Boltz-Flow raises the average success rate of $\pi_{0.5}$ from 94.65\% to 98.55\%, and subsequent noise-adapted fine-tuning raises it to 99.05\%. On LIBERO-Plus, it improves the seven-category average from 82.3\% to 85.9\%, an absolute gain of 3.6 percentage points (pp), with improvements in all 7 categories. On RoboTwin 2.0, it increases success rates from 78.68\% to 80.82\% under clean conditions and from 54.54\% to 56.92\% under randomized conditions. CIM-based training also converges in fewer iterations than SA-based training on LIBERO. Moreover, the gains stack with action-to-action initialization~\cite{jia2026a2a} and transfer to the SmolVLA backbone~\cite{smolvla2025}.

\section{Related Work}
\paragraph{Source Initialization for Flow Policies.}
\label{sec:initialization_strategies}

Flow-matching VLAs generate actions by transporting a source distribution to the action distribution, and the initial noise sets the boundary condition of this finite-step integration~\cite{lipman2023flow}. In general generative modeling, prior work improves the starting point by learning noise transformations, searching or selecting initial noise, and predicting context-dependent \textit{Gaussian} priors~\cite{scholz2025warm}. In robot action generation, existing initialization strategies fall into four categories. Unconditional initialization samples from an observation-independent standard \textit{Gaussian} and remains the default in $\pi_0$ and $\pi_{0.5}$~\cite{black2024pi0,physicalintelligence2025pi05}. Learning-based structured initialization builds a starting point with richer action structure from the task context~\cite{zhai2025vfp,du2026cfvla}. History-based initialization exploits temporal continuity and builds the next starting point from previous actions~\cite{jia2026a2a,jia2026mars}. Noise-space selection and adaptation search, filter, invert, or learn the starting point in the input space of a frozen policy~\cite{wagenmaker2025dsrl}. Unlike these strategies, Boltz-Flow learns a source that is conditioned on the observation, coupled across time steps and action dimensions, and able to hold several low-energy modes at once.

\paragraph{Quantum Computing for Energy-Based Models.}
Quantum annealers have been used to estimate the negative-phase statistics of BMs and to train hardware-embedded probabilistic graphical models~\cite{benedetti2017quantum}. CIMs encode spins with optical parametric oscillators and realize densely connected Ising optimization through time multiplexing and measurement-feedback coupling~\cite{inagaki2016coherent}. Noise-injected optoelectronic Ising machines further support approximate \textit{Boltzmann} sampling and restricted BM training~\cite{bohm2022noise}. Quantum BMs define energy models through quantum Gibbs states~\cite{amin2018quantum}. \textit{Boltzmann} priors for the latent variables of variational autoencoders (VAEs) have been implemented with quantum Monte Carlo, D-Wave annealers, and CIMs~\cite{khoshaman2018quantum,vinci2020path,wang2025quantum}. D-Wave-sampled \textit{Boltzmann} priors have also replaced random latent variables in generative adversarial networks (GANs)~\cite{wilson2021quantum}. These works target generic generative models and their training. In contrast, Boltz-Flow uses a CIM to search the source of a robot policy, which is, to our knowledge, the first use of a quantum computer for continuous robot action generation.

\section{Method}
\label{sec:method}

\subsection{Problem Formulation}
\label{sec:problem_notation}

Flow matching~\citep{lipman2023flow} is a representative generative paradigm for continuous-control VLAs~\citep{physicalintelligence2025pi05, black2024pi0}. Given the robot observation $o_t\in\mathcal O$ and language instruction $l\in\mathcal L$, where $\mathcal O$ and $\mathcal L$ denote the observation and instruction spaces, it integrates a conditional velocity field over flow time $\tau\in[0,1]$. The integration transports a source sample at $\tau=0$ into an action chunk of $H\in\mathbb{N}$ steps and $d_a\in\mathbb{N}$ action dimensions at $\tau=1$. Let $\boldsymbol{\mathcal G}_\theta$ denote the complete $K$-step numerical integration ($K\in\mathbb{N}$) of the learned velocity field with parameters $\theta$. Standard generation is written as
\begin{equation}
  \hat{\mathbf a}
  =
  \boldsymbol{\mathcal G}_\theta(o_t,l;\boldsymbol{\epsilon}),
  \qquad
  \boldsymbol{\epsilon}
  \sim
  \mathcal N(\mathbf 0,\mathbf I),
  \label{eq:noise_conditioned_policy}
\end{equation}
where the initial noise $\boldsymbol{\epsilon}\in\mathbb{R}^{H\times d_a}$ follows the  \textit{Gaussian} distribution and $\hat{\mathbf a}\in\mathbb{R}^{H\times d_a}$ is the predicted action chunk.

This noise is independent of the observation and instruction, and independent across time steps and control dimensions. However, for a fixed observation and a frozen policy, different noise samples produce action chunks whose errors with respect to the demonstration vary substantially. We therefore replace the fixed \textit{Gaussian} prior with a condition-dependent noise sampler $\boldsymbol{\mathcal S}_\psi$,
\begin{equation}
  \tilde{\boldsymbol{\epsilon}}
  =
  \boldsymbol{\mathcal S}_\psi(o_t,l;\boldsymbol{\xi}),
  \label{eq:conditional_noise_sampler}
\end{equation}
where $\psi$ denotes the sampler parameters, $\boldsymbol{\xi}\in\Xi$ represents its internal randomness with sample space $\Xi$, and $\tilde{\boldsymbol{\epsilon}}\in\mathbb{R}^{H\times d_a}$ is the generated noise. This noise replaces $\boldsymbol{\epsilon}$ in $\boldsymbol{\mathcal G}_\theta$, and the policy stays frozen. This formulation covers learned latent-noise policies, deterministic initializers, and energy-based samplers. Our goal is a sampler that assigns higher probability to noise that produces lower-error actions.

\begin{figure*}[t]
  \centering
  \includegraphics[width=1\textwidth]{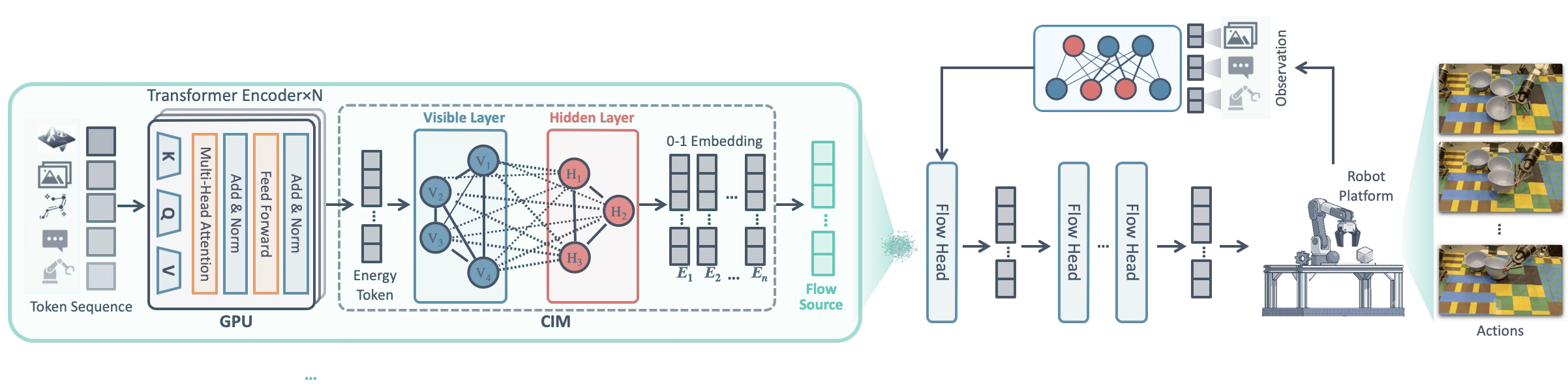}
  \caption{\textbf{Overview of Boltz-Flow.} \textbf{Right.} A conditioning
  module replaces the \textit{Gaussian} source of a frozen flow-matching
  policy. It maps the robot observation and instruction to a flow source,
  which the frozen flow heads integrate into an action chunk for the robot.
  \textbf{Left.} Inside the conditioning module, a Transformer encoder on the
  GPU turns the observation and instruction tokens into energy tokens, which
  clamp the visible layer of a conditional BM. The CIM searches the hidden
  layer for low-energy binary states. A midpoint-\textit{Gaussian}-quantile
  mapping then decodes these states (0-1 embedding) into the continuous flow
  source.}
  \label{fig:noise_optimization_architecture}
\end{figure*}

Boltz-Flow instantiates $\boldsymbol{\mathcal S}_\psi$ with a conditional BM, as illustrated in Fig.~\ref{fig:noise_optimization_architecture}. The observation and instruction define an energy over binary states, and Ising search returns low-energy states. A midpoint-\textit{Gaussian}-quantile mapping turns the selected state into continuous noise, from which the frozen policy generates the action chunk without any change to its velocity field. Sec.~\ref{sec:noise_optimization} details this source, and Sec.~\ref{sec:frozen_flow_labeling} describes how it is trained.

\subsection{Boltzmann Flow-Matching Policy}
\label{sec:noise_optimization}

\paragraph{Conditional Boltzmann Machine.}
Rather than regressing continuous noise directly, the BM models a distribution over binary states through its energy landscape. A multimodal Transformer encodes the observation and instruction into a fixed-length representation $\mathbf{z}=\mathbf{f}_\phi(o_t,l)\in\mathbb{R}^{V}$, where $\phi$ denotes the encoder parameters and $V\in\mathbb{N}$. We treat $\mathbf{z}$ as the visible condition nodes of the BM and let $\mathbf{h}\in\{0,1\}^{N_h}$ denote its $N_h\in\mathbb{N}$ binary hidden nodes. With the complete state $\mathbf{x}=[\mathbf{z};\mathbf{h}]\in\mathbb{R}^{V+N_h}$, the conditional energy is
\begin{equation}
  E_\psi(\mathbf{z},\mathbf{h})
  =
  -\mathbf{x}^{\mathsf{T}}\mathbf{b}
  -\mathbf{x}^{\mathsf{T}}\mathbf{Q}\mathbf{x},
  \label{eq:conditional_bm_energy}
\end{equation}
where $\mathbf{Q}\in\mathbb{R}^{(V+N_h)\times(V+N_h)}$ is the quadratic coupling matrix and $\mathbf{b}\in\mathbb{R}^{V+N_h}$ is the linear bias. We write $\psi=\{\phi,\mathbf{Q},\mathbf{b}\}$ for all learnable parameters. With the visible nodes clamped to $\mathbf{z}$, the energy induces the conditional \textit{Boltzmann} distribution
\begin{equation}
  \hat{\mathbf h}
  \sim
  p_\psi(\mathbf h\mid\mathbf z)
  =
  \frac{\exp[-E_\psi(\mathbf z,\mathbf h)]}
       {Z_\psi(\mathbf z)},
  \label{eq:conditional_boltzmann_distribution}
\end{equation}
where $\hat{\mathbf h}\in\{0,1\}^{N_h}$ is a sampled hidden state and the partition function $Z_\psi(\mathbf z)>0$ normalizes over all binary states. Lower-energy states thus receive exponentially higher probability. Because the landscape can hold several separated low-energy basins, sampling can return states from distinct modes. The source therefore stays diverse while adapting to the current observation and instruction.

\paragraph{Ising-Based Sampling.}
For a fixed $\mathbf{z}$, collecting all terms that depend on $\mathbf{h}$ gives the binary quadratic energy
\begin{equation}
  E_{\mathbf z}(\mathbf{h})
  =
  C(\mathbf z)
  +\sum_{a=1}^{N_h}\alpha_{a}(\mathbf z)\,h_a
  +\sum_{1\leq a<a'\leq N_h}
  \beta_{a,a'}h_ah_{a'},
  \label{eq:conditional_bqm}
\end{equation}
where $C(\mathbf z)\in\mathbb{R}$ does not depend on $\mathbf{h}$, $\alpha_{a}(\mathbf z)\in\mathbb{R}$ is the condition-dependent linear coefficient of hidden node $a$, and $\beta_{a,a'}\in\mathbb{R}$ is the coupling between hidden nodes $a,a'\in\{1,\ldots,N_h\}$. The standard affine bit-to-spin map $s_a=2h_a-1\in\{-1,+1\}$ turns Eq.~\eqref{eq:conditional_bqm} into an equivalent Ising problem with local fields and pairwise spin couplings. This conversion preserves the ordering of hidden states up to an additive constant and any positive rescaling that the hardware requires. Finding the lowest-energy state is NP-hard. We therefore use the CIM~\cite{wei2026cim}, a quantum-optical computer for Ising optimization. It carries computation on coherent photonic states and efficiently explores low-energy configurations of large Ising systems. The CIM returns one or more low-energy spin configurations, which are mapped back to binary states $\hat{\mathbf h}$. The full coefficient transformation and the augmented-matrix construction are given in the supplementary material.

\paragraph{From Binary States to Noise.}
The BM uses
\begin{equation}
  N_h = H d_e B_q
  \label{eq:bm_binary_dimension}
\end{equation}
binary variables, where $d_e\in\mathbb{N}$, $d_e\leq d_a$, is the number of action dimensions modeled by the BM, and $B_q\in\mathbb{N}$ is the number of Gray-code bits per noise coordinate. Each group of $B_q$ bits is decoded into an integer $\kappa\in\{0,\ldots,2^{B_q}-1\}$ and mapped to the midpoint \textit{Gaussian} quantile $\Phi^{-1}((\kappa+1/2)/2^{B_q})$, where $\Phi^{-1}$ is the inverse cumulative distribution function of the standard \textit{Gaussian}. The remaining $d_a-d_e$ dimensions keep independent standard \textit{Gaussian} values. The resulting noise $\boldsymbol{\epsilon}(\hat{\mathbf h})\in\mathbb{R}^{H\times d_a}$ is injected once as the initial state of $\boldsymbol{\mathcal G}_\theta$. Neither the input of the velocity field nor the integration procedure changes.

\paragraph{CPU--GPU--CIM Execution.}
The pipeline splits the work across three processors. The GPU runs the Transformer that produces $\mathbf{z}$, evaluates the differentiable BM energies during training, and generates the actions. The CPU runs the policy loop, exchanges sensor and action data with the environment, builds the Ising instance $\mathcal{I}$ from $\mathbf{z}$, calls the solver, and decodes the returned set of spin configurations $\mathcal{Y}\subseteq\{-1,+1\}^{N_h}$. The CIM receives only the Ising coefficients and never processes raw images, language instructions, or gradients. The resulting data flow is
\begin{equation}
\begin{aligned}
  o_t
  &\xrightarrow{\mathrm{GPU}}
  \mathbf{z}
  \xrightarrow{\mathrm{CPU}}
  \mathcal{I}
  \xrightarrow{\mathrm{CIM}}
  \mathcal{Y},
  \\
  \mathcal{Y}
  &\xrightarrow{\mathrm{CPU}}
  \hat{\mathbf{h}}
  \xrightarrow{\mathrm{GPU}}
  \hat{\mathbf{a}}
  \xrightarrow{\mathrm{CPU/GPU}}
  \mathrm{environment}.
\end{aligned}
\label{eq:hybrid_dataflow}
\end{equation}
Only the combinatorial search thus runs on the CIM, while perception, learning, action generation, and environment interaction stay on the classical CPU--GPU system.

\subsection{Training Objective}
\label{sec:frozen_flow_labeling}

We train the BM with supervision from the frozen policy. For each tuple $(o_t,l,\mathbf a^\star)$ drawn from the demonstration dataset $\mathcal D$, where $\mathbf a^\star\in\mathbb{R}^{H\times d_a}$ is the expert action chunk, we build a candidate set $\{\mathbf h_k\}_{k=1}^{M}\subset\{0,1\}^{N_h}$ of $M\in\mathbb{N}$ states. It mixes low-energy states proposed by the BM with random states. Each state is mapped to noise for the frozen generator, $\hat{\mathbf a}_k=\boldsymbol{\mathcal G}_\theta(o_t,l;\boldsymbol{\epsilon}(\mathbf h_k))\in\mathbb{R}^{H\times d_a}$, with $\theta$ fixed.

We score each candidate by its error on the $d_e$ action dimensions modeled by the BM, $e_k=(Hd_e)^{-1}\|\hat{\mathbf a}_{k,e}-\mathbf a^\star_e\|_F^2\in\mathbb{R}_{\geq0}$, where the subscript $e$ selects these dimensions so that $\hat{\mathbf a}_{k,e},\mathbf a^\star_e\in\mathbb{R}^{H\times d_e}$. The error is a stop-gradient signal and does not backpropagate through $\boldsymbol{\mathcal G}_\theta$. To remove scale differences across examples, we standardize the errors within each candidate set as $\tilde e_k=(e_k-\mu_e)/(\sigma_e+\nu)$, where $\mu_e\in\mathbb{R}$ and $\sigma_e\in\mathbb{R}_{\geq0}$ are their mean and standard deviation and $\nu>0$ ensures numerical stability. The standardized errors define target weights, and the energies define model probabilities over the same set,
\begin{equation}
  \begin{aligned}
    w_k
    &=
    \frac{\exp(-\tilde e_k/T_e)}
         {\sum_{j=1}^{M}\exp(-\tilde e_j/T_e)},\\
    p_{\psi,k}
    &=
    \frac{\exp[-E_\psi(\mathbf z,\mathbf h_k)]}
         {\sum_{j=1}^{M}\exp[-E_\psi(\mathbf z,\mathbf h_j)]},
  \end{aligned}
  \label{eq:target_energy_distribution}
\end{equation}
where $w_k,p_{\psi,k}\in(0,1)$ and the temperature $T_e>0$ controls how concentrated the target weights are. A listwise cross-entropy aligns the two distributions,
\begin{equation}
  \mathcal L_{\mathrm{noise}}
  =
  \mathbb E
  \left[
    -\sum_{k=1}^{M}
    w_k\log p_{\psi,k}
  \right]
  +
  \lambda\mathcal R(\psi),
  \label{eq:noise_loss}
\end{equation}
where the expectation runs over training examples and candidate sets, and $\mathcal R(\psi)\in\mathbb{R}_{\geq0}$ is a regularizer with coefficient $\lambda\geq0$. The BM thus learns to assign lower energy to states whose noise yields better actions after frozen integration, without updating the base policy.

\section{Experiments}
\label{sec:boltz_flow_experiments}

We evaluate Boltz-Flow on LIBERO~\cite{liu_zhu_NeurIPS2023}, LIBERO-Plus~\cite{Fei_2026_CVPR}, and RoboTwin 2.0, comparing it with strong imitation learning baselines and state-of-the-art source initialization methods. Controlled ablations examine the effects of the initial source distribution and full-parameter fine-tuning. We then compare task success rates across hardware solver backends used for training and inference, analyze training convergence efficiency, and study the impact of Gray-code precision. We further assess cross-backbone transfer to SmolVLA~\cite{smolvla2025} and composability with Action-to-Action~\cite{jia2026a2a}. Finally, we conduct real-world robot experiments on five tasks adapted from RoboTwin 2.0 to compare the manipulation performance of Boltz-Flow and OpenPI.

\subsection{Experimental Setup}

\paragraph{Benchmarks and metrics.}
We evaluate Boltz-Flow on standard LIBERO, LIBERO-Plus, and RoboTwin2.0. LIBERO has 4 suites, Spatial, Object, Goal, and Long, with 10 tasks per suite. We follow the standard training protocol and conduct 50 evaluation rollouts per task, totaling 2,000 rollouts per method. LIBERO-Plus applies 7 controlled perturbations, Camera, Robot, Language, Light, Background, Noise, and Layout, to the same 40 tasks and contains 10,030 evaluation instances. LIBERO-Plus is used exclusively for testing. We report the success rate of each suite and their mean. RoboTwin 2.0~\cite{chen2025robotwin20} contains 50 bimanual tasks. We conduct 100 rollouts per task in each of the Clean and Random settings and report the mean task success rate(SR).

\paragraph{Policy backbones and evaluation protocols.}
Our primary backbone is the OpenPI implementation of the $\pi_{0.5}$ flow-matching policy~\cite{physicalintelligence2025pi05}. We distinguish two settings. \emph{Boltz-Flow} trains the conditional energy-based noise module while keeping the policy frozen. \emph{Boltz-Flow + noise-adapted full fine-tuning} subsequently updates all $\pi_{0.5}$ parameters to adapt the velocity field to the learned source distribution. Controlled comparisons use the same action normalization and 10-step \textit{Euler} integration budget.
The execution horizon $r\in\{1,\ldots,H\}$ denotes the number of predicted actions executed before the policy is queried again. We set both the prediction horizon $H$ and the execution horizon $r$ to 10 for the controlled comparisons below. To accommodate the CIM's capacity of 1,000 spin variables, we encode each noise component using 7-bit Gray codes for bimanual RoboTwin 2.0 tasks ($d_a=14$) and 14-bit Gray codes for single-arm LIBERO tasks ($d_a=7$). Both settings use $N_h=980$ binary variables. FT denotes subsequent noise-adapted full fine-tuning of the corresponding policy.
Given limited access to CIM resources, we use the CIM as the training-time energy solver only on LIBERO. The resulting models are evaluated on LIBERO with both CIM and SA inference, and on LIBERO-Plus with SA inference. RoboTwin 2.0 experiments use SA for energy minimization during both training and inference. The supplementary material reports additional solver diagnostics.
We reproduce LEAP, VITA, Fast-Flow, and A2A with the OpenPI $\pi_{0.5}$ backbone, training each method for 30,000 optimization steps with a global batch size of 32. LEAP updates the base policy, whereas VITA, Fast-Flow, and A2A keep it frozen.

\subsection{Benchmark Results}


\begin{table}[t]
\centering
\renewcommand{\arraystretch}{0.82}
\caption{LIBERO success rates (\%). The upper block reports prior results;
the lower block reports our OpenPI evaluations. Entries are ordered by
the four-suite average, with $\pi_{0.5}$ first in the lower block.
``(CIM)'' denotes CIM training with SA inference.
FT denotes noise-adapted full fine-tuning; dashes indicate unreported results.
$\dagger$ denotes our reproduced $\pi_{0.5}$ baseline.}
\label{tab:boltz_flow_libero_main}
\small
\setlength{\tabcolsep}{0.8mm}
\begin{tabular}{@{}lccccc@{}}
\toprule
Method & Spa. & Obj. & Goal & Long & Avg. \\
\midrule
Diffusion Policy~\cite{chi2023diffusion} & 78.3 & 92.5 & 68.3 & 50.5 & 72.4 \\
Octo~\cite{octo_2024} & 78.9 & 85.7 & 84.6 & 51.1 & 75.1 \\
OpenVLA~\cite{kim2025openvla} & 84.7 & 88.4 & 79.2 & 53.7 & 76.5 \\
SpatialVLA~\cite{qu2025spatialvla} & 88.2 & 89.9 & 78.6 & 55.5 & 78.1 \\
WorldVLA~\cite{worldvla2025} & 87.6 & 96.2 & 83.4 & 60.0 & 81.8 \\
CoT-VLA~\cite{zhao2025cotvla} & 87.5 & 91.6 & 87.6 & 69.0 & 83.9 \\
$\pi_0$-FAST~\cite{pertsch2025fast} & 96.4 & 96.8 & 88.6 & 60.2 & 85.5 \\
SmolVLA-2.25B~\cite{smolvla2025} & 93.0 & 94.0 & 91.0 & 77.0 & 88.8 \\
Qwen-VLA Base~\cite{qwen2026qwenvla} & -- & -- & -- & -- & 90.8 \\
CogACT~\cite{li2024cogact} & 97.2 & 98.0 & 90.2 & 88.8 & 93.6 \\
$\pi_0$~\cite{black2024pi0} & 96.8 & 98.8 & 95.8 & 85.2 & 94.2 \\
UniVLA~\cite{wang2025univla} & 95.4 & 98.8 & 93.6 & 94.0 & 95.5 \\
Discrete Diffusion VLA~\cite{liang2025discretediffusion} & 97.2 & 99.4 & 96.8 & 92.2 & 96.4 \\
StarVLA~\cite{starvla2026} & 97.8 & 98.6 & 96.2 & 93.8 & 96.6 \\
MemoryVLA~\cite{shi2025memoryvla} & 98.4 & 98.4 & 96.4 & 93.4 & 96.7 \\
GR00T-N1.6~\cite{nvidia2025grootn1} & 97.7 & 98.5 & 97.5 & 94.4 & 97.0 \\
OpenVLA-OFT~\cite{kim2025openvlaoft} & 97.6 & 98.4 & 97.9 & 94.5 & 97.1 \\
VITA-VLA~\cite{dong2025vitavla} & 98.0 & 99.8 & 97.9 & 93.5 & 97.3 \\
$\pi_{0.5}$ + GridS$_{16}$~\cite{feng2026gridsampler} & 98.6 & 98.8 & 98.4 & 95.2 & 97.8 \\
Qwen-VLA Instruct~\cite{qwen2026qwenvla} & -- & -- & -- & -- & 97.9 \\
$\pi_{0.5}$ + Flow SDE~\cite{chen2025pirl} & 99.6 & 100.0 & 98.8 & 93.0 & 97.9 \\
$\pi_{0.5}$ + Flow-Noise~\cite{chen2025pirl} & 99.6 & 100.0 & 99.6 & 94.0 & 98.3 \\
\midrule
\color{darkgray} $\pi_{0.5}^{\dagger}$ (Pytorch)~\cite{physicalintelligence2025pi05} & 96.4 & 98.8 & 95.8 & 87.6 & 94.7 \\
LEAP~\cite{dai2026leap} & 98.6 & 98.8 & 98.0 & 92.2 & 96.9 \\
Fast-Flow~\cite{sochopoulos2025cot} & 98.6 & 99.0 & 98.0 & 92.6 & 97.1 \\
VITA~\cite{gao2026vita} & 99.4 & 98.6 & 97.8 & 93.0 & 97.2 \\
A2A~\cite{jia2026a2a} & 98.6 & 99.4 & 97.4 & 93.8 & 97.3 \\
\textbf{Boltz-Flow (CIM)} & \textbf{99.4} & \textbf{99.8} & \textbf{98.6} & \textbf{96.4} & \textbf{98.6} \\
\textbf{Boltz-Flow + FT (CIM) } & \textbf{99.8} & \textbf{99.6} & \textbf{99.2} & \textbf{97.6} & \textbf{99.1} \\
\bottomrule
\end{tabular}
\end{table}

\noindent\textbf{LIBERO.}
Table~\ref{tab:boltz_flow_libero_main} separates reported results from
our OpenPI evaluations. Frozen Boltz-Flow achieves 98.6\% average success,
surpassing the strong baselines and prior state-of-the-art methods for
source-distribution optimization, with a 1.3 percentage-point
lead over A2A. It improves over OpenPI (94.7\% average) on all four suites,
with the largest gain on Long-Horizon (87.6\% to 96.4\%).
Noise-adapted fine-tuning further raises average success to 99.1\%.

\begin{table*}[t]
\centering
\renewcommand{\arraystretch}{0.9}
\caption{LIBERO-Plus success rates (\%) under seven perturbations.
Overall is the arithmetic mean across categories.
The upper block reports prior results; the lower block reports our OpenPI
evaluations. Baselines are ordered by Overall, with $\pi_{0.5}$ first
and Boltz-Flow variants last in the lower block.
``(CIM)'' denotes CIM training with SA inference.
LIBERO-Plus is used only for evaluation.
FT denotes noise-adapted full fine-tuning.
$\dagger$ denotes our reproduced $\pi_{0.5}$ baseline.}
\label{tab:boltz_flow_libero_plus}
\small
\setlength{\tabcolsep}{3.6pt}
\begin{tabular}{@{}lcccccccc@{}}
\toprule
Method & Camera & Robot & Language & Light & Background & Noise & Layout & Overall \\
\midrule
OpenVLA~\cite{kim2025openvla} & 0.8 & 3.5 & 23.0 & 8.1 & 34.8 & 15.2 & 28.5 & 16.3 \\
WorldVLA~\cite{worldvla2025} & 0.1 & 27.9 & 41.6 & 43.7 & 17.1 & 10.9 & 38.0 & 25.6 \\
NORA~\cite{declare2025nora} & 2.2 & 37.0 & 65.1 & 45.7 & 58.6 & 12.8 & 62.1 & 40.5 \\
UniVLA~\cite{wang2025univla} & 1.8 & 46.2 & 69.6 & 69.0 & 81.0 & 21.2 & 31.9 & 45.8 \\
$\pi_0$~\cite{black2024pi0} & 13.8 & 6.0 & 58.8 & 85.0 & 81.4 & 79.0 & 68.9 & 56.1 \\
$\pi_0$-FAST~\cite{pertsch2025fast} & 65.1 & 21.6 & 61.0 & 73.2 & 73.2 & 74.4 & 68.8 & 62.5 \\
OpenVLA-OFT~\cite{kim2025openvlaoft} & 56.4 & 31.9 & 79.5 & 88.7 & 93.3 & 75.8 & 74.2 & 71.4 \\
X-VLA~\cite{2toinf2025xvla} & 23.4 & 89.7 & 75.7 & 88.2 & 96.0 & 62.7 & 71.8 & 72.5 \\
MemoryVLA++~\cite{shi2026memoryvlapp} & 36.4 & 68.9 & 88.7 & 93.8 & 90.6 & 63.5 & 83.8 & 75.1 \\
VLA-JEPA~\cite{sun2026vlajepa} & 63.3 & 67.1 & 85.4 & 95.6 & 93.6 & 66.3 & 85.1 & 79.5 \\
PokeVLA~\cite{zheng2026pokevla} & 84.7 & 46.1 & 84.8 & 94.6 & 82.6 & 89.8 & 77.2 & 80.0 \\
$\pi_{0.5}$ + GridS$_{32}$~\cite{feng2026gridsampler} & 76.3 & 67.1 & 76.9 & 95.5 & 92.8 & 90.7 & 82.7 & 83.1 \\
ACoT-VLA~\cite{zhong2026acotvla} & 91.2 & 62.5 & 80.3 & 95.1 & 91.5 & 88.3 & 84.9 & 84.8 \\
\midrule
\color{darkgray} $\pi_{0.5}^{\dagger}$ (Pytorch)~\cite{physicalintelligence2025pi05} & 69.9 & 73.4 & 80.6 & 91.9 & 92.6 & 85.0 & 82.6 & 82.3 \\
Fast-Flow~\cite{sochopoulos2025cot} & 71.0 & 75.8 & 85.2 & 96.3 & 95.7 & 86.8 & 85.8 & 85.2 \\
A2A~\cite{jia2026a2a} & 79.2 & 68.7 & 85.9 & 95.2 & 94.5 & 91.9 & 85.5 & 85.8 \\
VITA~\cite{gao2026vita} & 68.2 & 79.2 & 86.9 & 95.7 & 96.7 & 89.3 & 86.8 & 86.1 \\
LEAP~\cite{dai2026leap} & 73.9 & 77.5 & 86.9 & 96.0 & 98.2 & 92.8 & 87.4 & 87.5 \\
\textbf{Boltz-Flow (CIM)} & 71.4 & 77.1 & 85.0 & 97.4 & 97.9 & 85.4 & 87.0 & 85.9 \\
\textbf{Boltz-Flow + FT (CIM)} & 72.2 & 79.5 & 86.5 & 97.2 & 97.1 & 90.9 & 87.9 & 87.3 \\
\bottomrule
\end{tabular}
\end{table*}

\noindent\textbf{LIBERO-Plus.}
Table~\ref{tab:boltz_flow_libero_plus} shows that frozen Boltz-Flow
raises Overall success from 82.3\% to 85.9\%, improving all seven categories.
The largest gains are in Light (+5.5) and Background
(+5.3), followed by Language and Layout (+4.4 each).
Noise-adapted fine-tuning reaches 87.3\%, improving five categories over
frozen Boltz-Flow, with the largest gain in Noise (+5.5).
Light decreases from 97.4\% to 97.2\%, and Background from 97.9\% to 97.1\%.
The selected self-evaluated baselines achieve 87.5\% (LEAP), 86.1\%
(VITA), 85.8\% (A2A), and 85.2\% (Fast-Flow) Overall.

\noindent\textbf{RoboTwin.}
RoboTwin 2.0~\cite{chen2025robotwin20} is evaluated under Clean and Random settings. 
Clean uses the standard configuration. Random applies the full set of benchmark randomizations, including color changes, variations in natural-language instructions, and distractor objects.
Table~\ref{tab:boltz_flow_robotwin} separates reported
results from our OpenPI evaluations. The reported baselines use 50 Clean
demonstrations per task and are evaluated on 50 tasks with 100 trials per task
in each setting~\cite{robotwin2026leaderboard,shou2026halo}.
VITA freezes the base policy, whereas LEAP allows it to be updated. Boltz-Flow improves both settings over OpenPI. With noise-adapted
fine-tuning, it reaches 81.28\% and 57.56\% on Clean and Random, respectively,
the highest rates among the listed methods.

\begin{table}[t]
\centering
\renewcommand{\arraystretch}{0.9}
\caption{RoboTwin 2.0 success rates (\%). The upper block reports prior
results from the accepted paper; the lower block reports our OpenPI evaluations.
Entries are ordered by Random success, with $\pi_{0.5}$ first in the lower block.
Clean uses the standard setting; Random includes all randomizations and distractors.
FT denotes noise-adapted full fine-tuning.$\dagger$ denotes our  $\pi_{0.5}$ baseline.}
\label{tab:boltz_flow_robotwin}
\small
\setlength{\tabcolsep}{6pt}
\begin{tabular}{@{}lcc@{}}
\toprule
Method & Clean & Random \\
\midrule
Fast-WAM~\cite{yuan2026fastwam} & 77.76 & 1.92 \\
Evo-1~\cite{lin2026evo1} & 65.70 & 12.50 \\
EventVLA~\cite{yang2026eventvla} & 65.58 & 15.74 \\
X-VLA~\cite{2toinf2025xvla} & 67.96 & 20.95 \\
Spatial Forcing~\cite{li2026spatialforcing} & 77.18 & 26.74 \\
NoTVLA~\cite{huang2026notvla} & 63.28 & 28.40 \\
HALO~\cite{shou2026halo} & 80.70 & 37.60 \\
\midrule
\color{darkgray} $\pi_{0.5}^{\dagger}$ (Pytorch)~\cite{physicalintelligence2025pi05} & 78.68 & 54.54 \\
VITA (frozen)~\cite{gao2026vita} & 57.08 & 42.08 \\
LEAP (unfrozen)~\cite{dai2026leap} & 73.98 & 53.88 \\
Fast-Flow~\cite{sochopoulos2025cot} & 78.36 & 54.50 \\
A2A~\cite{jia2026a2a} & 78.30 & 55.72 \\
\textbf{Boltz-Flow} & \textbf{80.82} & \textbf{56.92} \\
\textbf{Boltz-Flow + adapted FT} & \textbf{81.28} & \textbf{57.56} \\
\bottomrule
\end{tabular}
\end{table}

\subsection{Ablation Studies}

\subsubsection{Initial Source Distribution}

\begin{table}[t]
\centering
\renewcommand{\arraystretch}{0.82}
\caption{Ablation of the initial source distribution. All entries
are success rates (\%); the base policy is frozen in all three settings.}
\label{tab:boltz_flow_source_ablation}
\small
\setlength{\tabcolsep}{1mm}
\begin{tabular}{@{}lccccc@{}}
\toprule
Method & Spa. & Obj. & Goal & Long & Avg. \\
\midrule
\color{darkgray} $\pi_{0.5}$~\cite{physicalintelligence2025pi05} & 96.4 & 98.8 & 95.8 & 87.6 & 94.65 \\
$\pi_{0.5}$ + VAE noise & 97.8 & 98.8 & 96.8 & 92.8 & 96.55 \\
$\pi_{0.5}$ + Boltz-Flow & 99.4 & 98.6 & 98.2 & 96.2 & 98.10 \\
\bottomrule
\end{tabular}

\end{table}

Table~\ref{tab:boltz_flow_source_ablation} isolates the effect of the initial
source distribution under the same OpenPI backbone. A VAE source raises average success from 94.65\% to 96.55\%, showing
that learning an observation-dependent initialization is already useful.
Boltz-Flow reaches 98.10\%, improving over standard Gaussian noise by 3.45
points and over VAE noise by 1.55 points. The difference is most pronounced on
Long-Horizon: VAE noise reaches 92.8\%, whereas Boltz-Flow reaches 96.2\%, a
further gain of 3.4 points.

This ablation locates the key benefit in the structure of the source rather
than in learned noise alone. Under partial task constraints, the
observation-conditioned Gibbs--Boltzmann distribution can reweight several
low-energy initial configurations while its pairwise energy terms capture
dependencies across time steps and control dimensions. The especially large
Long-Horizon gain is consistent with this joint representation: multi-stage
tasks place stronger demands on the coordination of position, orientation, and
gripper control throughout an action block.

\subsubsection{Training Strategy and Noise Adaptation}

\begin{table}[t]
\centering
\renewcommand{\arraystretch}{0.9}
\caption{Ablation of policy adaptation. Full-parameter FT uses standard Gaussian noise, whereas
noise-adapted FT adapts the velocity field to the Boltzmann source. All entries are
success rates.}
\label{tab:boltz_flow_replanning_ablation}
\small
\setlength{\tabcolsep}{1mm}
\begin{tabular}{@{}lcccccc@{}}
\toprule
$\pi_{0.5}$ variant & Spa. & Obj. & Goal & Long & Avg. \\
\midrule
\color{darkgray} $\pi_{0.5}$~\cite{physicalintelligence2025pi05} & 96.4 & 98.8 & 95.8 & 87.6 & 94.65 \\
+ full-parameter FT & 98.0 & 99.2 & 97.8 & 96.0 & 97.75 \\
+ Boltz-Flow & 99.4 & 98.6 & 98.2 & 96.2 & 98.10 \\
+ Boltz-Flow + adapted FT & 99.2 & 99.8 & 98.4 & 97.4 & 98.70 \\
\bottomrule
\end{tabular}

\end{table}

Table~\ref{tab:boltz_flow_replanning_ablation} compares Boltz-Flow with
full-parameter fine-tuning. Frozen Boltz-Flow reaches 98.10\%,
outperforming the OpenPI baseline by
3.45 pp and full-parameter fine-tuning by 0.35 pp. A conditioned source thus improves access to control priors in the frozen VLA.
Noise-adapted fine-tuning raises success from 98.10\% to 98.70\%. Object and Long-Horizon each
gain 1.2 pp, reaching 99.8\% and 97.4\%. 
Boltz-Flow improves the initial
source, and fine-tuning adapts the velocity field to it, making the stages complementary.

\subsubsection{Training Solvers and Efficiency}

\begin{figure*}[t]
    \centering
    \includegraphics[width=\textwidth]{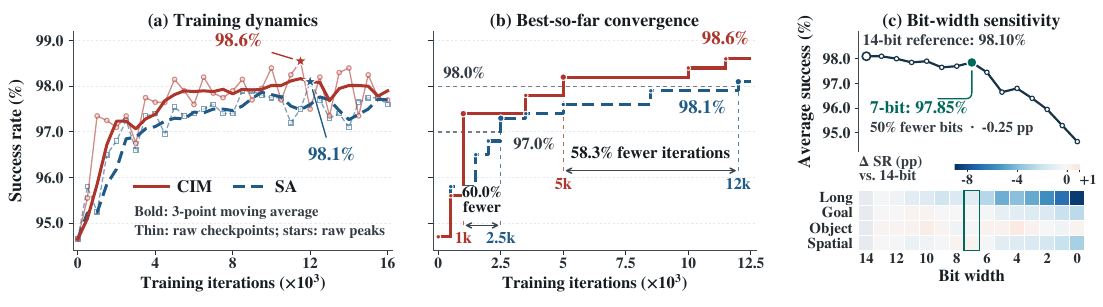}
    \caption{\textbf{Solver convergence and Gray-code precision on standard LIBERO.}
    Success rates are averaged equally over the Spatial, Object, Goal,
    and Long-Horizon suites.
    \textbf{(a)} Checkpoint success rates (thin curves) and three-point
    moving averages (thick curves) through 16,000 training iterations.
    Stars indicate the highest observed checkpoint success rate for each solver.
    \textbf{(b)} Best-so-far success rates through 12,500 training iterations.
    The marked iterations indicate when each solver reaches the 97.0\%
    and 98.0\% success thresholds; percentage annotations report the
    reduction in training iterations relative to SA.
    \textbf{(c)} Average success versus Gray-code bit width (top) and
    suite-level changes in percentage points relative to the 14-bit
    reference (bottom). ``Long'' denotes LIBERO-10.
    The highlighted 7-bit configuration halves the number of binary
    source variables with a 0.25 percentage-point decrease in average success.}
    \label{fig:convergence_precision}
\end{figure*}

\paragraph{Solver convergence.}
Figure~\ref{fig:convergence_precision}(a,b) compares the training
convergence of CIM and SA on standard LIBERO. Panel~(a) shows checkpoint
success rates and their three-point moving averages, with peak success
rates of 98.6\% for CIM and 98.1\% for SA. Panel~(b) reports the best
success rate attained up to each training iteration. CIM reaches the
97.0\% and 98.0\% success thresholds after 1,000 and 5,000 iterations,
respectively, compared with 2,500 and 12,000 iterations for SA.
These correspond to reductions of 60.0\% and 58.3\% in the number of
training iterations required to reach the respective thresholds.
The Train and Infer columns in
Tables~\ref{tab:boltz_flow_solver_ablation} and~\ref{tab:boltz_flow_solver_ablation_plus}
specify the training and inference solvers, respectively; the main benchmark
results use CIM for training and SA for inference on LIBERO and LIBERO-Plus,
and SA for both stages on RoboTwin 2.0.
On LIBERO, replacing SA training with CIM training raises average success
from 98.1\% to 98.6\% with the policy frozen, and from 98.7\% to 99.1\%
after noise-adapted fine-tuning. Using CIM at inference further raises
the frozen model's success to 99.0\%.
On LIBERO-Plus, CIM training raises Overall success from 85.4\% to 85.9\%
with the policy frozen, and from 86.8\% to 87.3\% after noise-adapted fine-tuning.

\begin{table}[t]
\centering
\renewcommand{\arraystretch}{0.9}
\caption{Solver ablation on LIBERO. All entries are success rates (\%).
The upper block uses a frozen policy; FT denotes subsequent noise-adapted
full-parameter fine-tuning.}
\label{tab:boltz_flow_solver_ablation}
\small
\setlength{\tabcolsep}{0.5mm}
\begin{tabular}{@{}lccccccc@{}}
\toprule
Method & Train & Infer & Spa. & Obj. & Goal & Long & Avg. \\
\midrule
Boltz-Flow & SA & SA & 99.4 & 98.6 & 98.2 & 96.2 & 98.1 \\
Boltz-Flow & CIM & SA & 99.4 & 99.8 & 98.6 & 96.4 & 98.6 \\
Boltz-Flow & CIM & CIM & 100.0 & 99.6 & 99.2 & 97.0 & 99.0 \\
\midrule
Boltz-Flow + FT & SA & SA & 99.2 & 99.8 & 98.4 & 97.4 & 98.7 \\
Boltz-Flow + FT & CIM & SA & 99.8 & 99.6 & 99.2 & 97.6 & 99.1 \\
\bottomrule
\end{tabular}
\end{table}

\begin{table*}[t]
\centering
\renewcommand{\arraystretch}{0.9}
\caption{Training-solver ablation on LIBERO-Plus. All entries are success
rates (\%); Overall is the arithmetic mean across the seven perturbation
categories. The upper block uses a frozen policy; FT denotes subsequent
noise-adapted full-parameter fine-tuning.}
\label{tab:boltz_flow_solver_ablation_plus}
\small
\setlength{\tabcolsep}{3.6pt}
\begin{tabular}{@{}lcccccccccc@{}}
\toprule
Method & Train & Infer & Camera & Robot & Language & Light & Background & Noise & Layout & Overall \\
\midrule
Boltz-Flow & SA & SA & 69.5 & 76.1 & 85.9 & 96.8 & 97.7 & 85.5 & 86.5 & 85.4 \\
Boltz-Flow & CIM & SA & 71.4 & 77.1 & 85.0 & 97.4 & 97.9 & 85.4 & 87.0 & 85.9 \\
\midrule
Boltz-Flow + FT & SA & SA & 71.4 & 78.0 & 87.1 & 96.8 & 97.9 & 89.7 & 87.0 & 86.8 \\
Boltz-Flow + FT & CIM & SA & 72.2 & 79.5 & 86.5 & 97.2 & 97.1 & 90.9 & 87.9 & 87.3 \\
\bottomrule
\end{tabular}
\end{table*}

\subsection{Parameter Studies}

\subsubsection{Gray-Code Precision}

Figure~\ref{fig:convergence_precision}(c) shows how success varies with
the number of Gray-code bits per noise coordinate. Mean success remains
close to the 14-bit reference down to 7 bits.
Reducing the width from 14 to 7 bits halves the number of binary source
variables in Eq.~\eqref{eq:bm_binary_dimension}, from 980 to 490 at fixed
$H=10$ and $d_e=7$. Mean success decreases from 98.10\% to 97.85\%,
a drop of 0.25 percentage points. More aggressive precision reduction
incurs larger losses, with the heatmap identifying Long-Horizon as the
most sensitive suite. This suggests that moderate source compression
preserves most of the policy's success, while finer source resolution
is more consequential for long-horizon tasks. The reduction
applies to the binary source representation; the frozen velocity field
is unchanged.


\subsection{Composability and Transfer Test}

\subsubsection{Transfer to SmolVLA}
\begin{table}[t]
\centering
\renewcommand{\arraystretch}{0.9}
\caption{Cross-backbone transfer to SmolVLA. The reproduced
SmolVLA row is the controlled baseline; the official result is included as an
external reference. All entries are success rates (\%).}
\label{tab:boltz_flow_smolvla}
\small
\setlength{\tabcolsep}{1mm}
\begin{tabular}{@{}lccccc@{}}
\toprule
SmolVLA variant~\cite{smolvla2025} & Spa. & Obj. & Goal & Long & Avg. \\
\midrule
\color{darkgray} Official & 90.0 & 96.0 & 92.0 & 71.0 & 87.25 \\
Reproduction & 87.2 & 97.4 & 92.4 & 87.2 & 91.05 \\
Reproduction + Boltz-Flow & 89.8 & 97.6 & 95.2 & 89.0 & 92.90 \\
\bottomrule
\end{tabular}

\end{table}

Using the reproduced SmolVLA as the controlled baseline,
Table~\ref{tab:boltz_flow_smolvla} shows that Boltz-Flow raises success
from 91.05\% to 92.90\%. The gains are positive on every suite: +2.6 on
Spatial, +0.2 on Object, +2.8 on Goal, and +1.8 on Long-Horizon. This
cross-backbone improvement shows that Boltz-Flow is not tied to the particular
velocity-field implementation of OpenPI and can improve the source interface
of another flow-matching VLA.

\subsubsection{Compatibility with Action-to-Action}
\begin{table}[t]
\centering
\renewcommand{\arraystretch}{0.9}
\caption{Incremental compatibility with Action-to-Action (A2A). All entries are
success rates (\%).}
\label{tab:boltz_flow_Action-to-Action}
\small
\setlength{\tabcolsep}{1mm}
\begin{tabular}{@{}lcccccc@{}}
\toprule
$\pi_{0.5}$ variant & Spa. & Obj. & Goal & Long & Avg. \\
\midrule
\color{darkgray} $\pi_{0.5}$~\cite{physicalintelligence2025pi05} & 96.4 & 98.8 & 95.8 & 87.6 & 94.65 \\
+ A2A~\cite{jia2026a2a} & 98.6 & 99.4 & 97.4 & 93.8 & 97.30 \\
+ A2A + Boltz-Flow & 98.8 & 100.0 & 97.6 & 95.4 & 97.95 \\
\bottomrule
\end{tabular}

\end{table}

Table~\ref{tab:boltz_flow_Action-to-Action} tests whether Boltz-Flow can improve an
existing action-level warm start. The combination improves the
A2A average from 97.30\% to 97.95\% (+0.65), with Long-Horizon
increasing by 1.6 points to 95.4\% and positive increments on all four suites.
Thus, Boltz-Flow can can be used alongside an action warm start and continues to
improve the match between the initial distribution and the current task.

\subsection{Real-World Robot Experiments}

We compare the OpenPI implementation of $\pi_{0.5}$ with Boltz-Flow on 5 real-world robot tasks: placing medicine into a tray using both arms, stacking three bowls, handing over a block between the two arms, placing a can next to a pot, and scanning a barcode. We collect 200 demonstration trajectories per task, totaling 1,000 demonstrations. Each model is evaluated over 10 trials per task, totaling 50 trials per model. Table~\ref{tab:boltz_flow_real_world} reports per-task and mean success rates, together with Boltz-Flow's gains over OpenPI. 
Boltz-Flow increases the overall success rate from 46\% to 62\%, an absolute gain of 16 percentage points. It improves all 5 tasks. The largest gains occur in can placement (+30 pp). These improvements across multiple tasks support the practical benefit of learned source initialization for real-world manipulation.

\vspace{-1.0em}
\begin{table}[t]
\centering
\caption{Real-world success rates. Parentheses show successful/total trials.
Gains are absolute improvements over OpenPI in percentage points (pp).}
\label{tab:boltz_flow_real_world}
\small
\setlength{\tabcolsep}{3pt}
\renewcommand{\arraystretch}{1.05}
\resizebox{\linewidth}{!}{%
\begin{tabular}{@{}lccc@{}}
\toprule
Task & $\pi_{0.5}$ (OpenPI) & Boltz-Flow & Gain (pp) \\
\midrule
Stack three bowls
& 90\% (9/10) & 100\% (10/10) & +10 \\
Bimanual block handover
& 40\% (4/10) & 50\% (5/10) & +10 \\
Place can next to pot
& 50\% (5/10) & 80\% (8/10) & +30 \\
Scan barcode
& 0\% (0/10) & 10\% (1/10) & +10 \\
Place medicine in tray
& 50\% (5/10) & 70\% (7/10) & +20 \\
\midrule
\textbf{Overall}
& \textbf{46\% (23/50)}
& \textbf{62\% (31/50)}
& \textbf{+16} \\
\bottomrule
\end{tabular}%
}
\end{table}

\section{Conclusion}

Boltz-Flow learns an observation- and instruction-conditioned Boltzmann source for flow-matching policies. A listwise objective from offline demonstrations aligns source energy with action quality, enabling Ising optimization to select the initial state for a single flow integration. With $\pi_{0.5}$ frozen, Boltz-Flow improves performance on LIBERO, LIBERO-Plus, and RoboTwin 2.0, with particularly strong gains on LIBERO-Long. Controlled ablations support the effectiveness of source learning and its complementarity with noise-adapted fine-tuning. These gains extend to SmolVLA, compose with action-to-action initialization, and persist across five real-world manipulation tasks. On LIBERO, the physical CIM benefits both training and inference: compared with GPU-based SA, CIM-assisted training achieves higher final policy success and reaches matched success thresholds with 58.3\%–60.0\% fewer training iterations, while CIM inference further improves task success. These results demonstrate that source-distribution learning improves pretrained policies while preserving their standard flow integration, and that CIM-based energy optimization further enhances training convergence and inference performance.

\section*{LLM Usage Statement}
OpenAI Codex is used to assist coding and implementation, organizing and analyzing the experiment results, and refining the wordings and the figures of the paper. The authors confirm to take full responsibility for all submitted content, regardless of any LLM or AI-tool use.

{\small
\bibliographystyle{ieeenat_fullname}
\bibliography{references}

@inproceedings{liu_zhu_NeurIPS2023,
  title={{LIBERO}: Benchmarking knowledge transfer for lifelong robot learning},
  author={Liu, Bo and Zhu, Yifeng and Gao, Chongkai and Feng, Yihao and Liu, Qiang and Zhu, Yuke and Stone, Peter},
  booktitle={Proceedings of Advances in Neural Information Processing Systems},
  year={2023}
}

@inproceedings{zitkovich2023rt,
  title={{RT-2}: Vision-language-action models transfer web knowledge to robotic control},
  author={Zitkovich, Brianna and Yu, Tianhe and Xu, Sichun and Xu, Peng and Xiao, Ted and Xia, Fei and Wu, Jialin and Wohlhart, Paul and Welker, Stefan and Wahid, Ayzaan and others},
  booktitle={Proceedings of the Conference on Robot Learning},
  year={2023}
}

@inproceedings{kim2025openvla,
  title={{OpenVLA}: An open-source vision-language-action model},
  author={Kim, Moo Jin and Pertsch, Karl and Karamcheti, Siddharth and Xiao, Ted and Balakrishna, Ashwin and Nair, Suraj and Rafailov, Rafael and Foster, Ethan P. and Sanketi, Pannag R. and Vuong, Quan and Kollar, Thomas and Burchfiel, Benjamin and Tedrake, Russ and Sadigh, Dorsa and Levine, Sergey and Liang, Percy and Finn, Chelsea},
  booktitle={Proceedings of the Conference on Robot Learning},
  year={2025}
}

@article{kim2025openvlaoft,
  title={Fine-tuning vision-language-action models: Optimizing speed and success},
  author={Kim, Moo Jin and Finn, Chelsea and Liang, Percy},
  journal={arXiv preprint arXiv:2502.19645},
  year={2025}
}

@inproceedings{octo_2024,
  title={Octo: An open-source generalist robot policy},
  author={Ghosh, Dibya and Walke, Homer and Pertsch, Karl and Black, Kevin and Mees, Oier and Dasari, Sudeep and Hejna, Joey and Xu, Charles and Luo, Jianlan and others},
  booktitle={Proceedings of Robotics: Science and Systems},
  year={2024}
}

@article{pertsch2025fast,
  title={{FAST}: Efficient action tokenization for vision-language-action models},
  author={Pertsch, Karl and Stachowicz, Kyle and Ichter, Brian and Driess, Danny and Nair, Suraj and Vuong, Quan and Mees, Oier and Finn, Chelsea and Levine, Sergey},
  journal={arXiv preprint arXiv:2501.09747},
  year={2025}
}

@article{worldvla2025,
  title={{WorldVLA}: Towards autoregressive action world model},
  author={Cen, Jun and Yu, Chaohui and Yuan, Hangjie and Jiang, Yuming and Huang, Siteng and Guo, Jiayan and Li, Xin and Song, Yibing and Luo, Hao and Wang, Fan and others},
  journal={arXiv preprint arXiv:2506.21539},
  year={2025}
}

@article{qu2025spatialvla,
  title={{SpatialVLA}: Exploring spatial representations for visual-language-action model},
  author={Qu, Delin and Song, Haoming and Chen, Qizhi and Yao, Yuanqi and Ye, Xinyi and Ding, Yan and Wang, Zhigang and Gu, JiaYuan and Zhao, Bin and Wang, Dong and others},
  journal={arXiv preprint arXiv:2501.15830},
  year={2025}
}

@inproceedings{zhao2025cotvla,
  title={{CoT-VLA}: Visual chain-of-thought reasoning for vision-language-action models},
  author={Zhao, Qingqing and Lu, Yao and Kim, Moo Jin and Fu, Zipeng and Zhang, Zhuoyang and Wu, Yecheng and Li, Zhaoshuo and Ma, Qianli and Han, Song and Finn, Chelsea and others},
  booktitle={Proceedings of the IEEE/CVF Conference on Computer Vision and Pattern Recognition},
  year={2025}
}

@article{li2024cogact,
  title={{CogACT}: A foundational vision-language-action model for synergizing cognition and action in robotic manipulation},
  author={Li, Qixiu and Liang, Yaobo and Wang, Zeyu and Luo, Lin and Chen, Xi and Liao, Mozheng and Wei, Fangyun and Deng, Yu and Xu, Sicheng and Zhang, Yizhong and others},
  journal={arXiv preprint arXiv:2411.19650},
  year={2024}
}

@article{wang2025univla,
  title={Unified vision-language-action model},
  author={Wang, Yuqi and Li, Xinghang and Wang, Wenxuan and Zhang, Junbo and Li, Yingyan and Chen, Yuntao and Wang, Xinlong and Zhang, Zhaoxiang},
  journal={arXiv preprint arXiv:2506.19850},
  year={2025}
}

@article{shi2025memoryvla,
  title={{MemoryVLA}: Perceptual-cognitive memory in vision-language-action models for robotic manipulation},
  author={Shi, Hao and Xie, Bin and Liu, Yingfei and Sun, Lin and Liu, Feng and Wang, Tiancai and Zhou, Erjin and Fan, Haoqiang and Zhang, Xiangyu and Huang, Gao},
  journal={arXiv preprint arXiv:2508.19236},
  year={2025}
}

@article{nvidia2025grootn1,
  title={{GR00T N1}: An open foundation model for generalist humanoid robots},
  author={Bjorck, Johan and Casta{\~n}eda, Fernando and Cherniadev, Nikita and Da, Xingye and Ding, Runyu and Fan, Linxi and Fang, Yu and Fox, Dieter and Hu, Fengyuan and Huang, Spencer and others},
  journal={arXiv preprint arXiv:2503.14734},
  year={2025}
}

@article{declare2025nora,
  title={{NORA}: A small open-sourced generalist vision language action model for embodied tasks},
  author={Hung, Chia-Yu and Sun, Qi and Hong, Pengfei and Zadeh, Amir and Li, Chuan and Tan, U-Xuan and Majumder, Navonil and Poria, Soujanya},
  journal={arXiv preprint arXiv:2504.19854},
  year={2025}
}

@article{2toinf2025xvla,
  title={X-vla: Soft-prompted transformer as scalable cross-embodiment vision-language-action model},
  author={Zheng, Jinliang and Li, Jianxiong and Wang, Zhihao and Liu, Dongxiu and Kang, Xirui and Feng, Yuchun and Zheng, Yinan and Zou, Jiayin and Chen, Yilun and Zeng, Jia and others},
  booktitle={International Conference on Learning Representations},
  volume={2026},
  pages={60580--60606},
  year={2026}
}

@inproceedings{lipman2023flow,
  title={Flow matching for generative modeling},
  author={Lipman, Yaron and Chen, Ricky T. Q. and Ben-Hamu, Heli and Nickel, Maximilian and Le, Matt},
  booktitle={Proceedings of the International Conference on Learning Representations},
  year={2023}
}

@inproceedings{chi2023diffusion,
  title={Diffusion policy: Visuomotor policy learning via action diffusion},
  author={Chi, Cheng and Feng, Siyuan and Du, Yilun and Xu, Zhenjia and Cousineau, Eric and Burchfiel, Benjamin C. M. and Song, Shuran},
  booktitle={Proceedings of Robotics: Science and Systems},
  year={2023}
}

@article{black2024pi0,
  title={$\pi_0$: A vision-language-action flow model for general robot control},
  author={Black, Kevin and Brown, Noah and Driess, Danny and Esmail, Adnan and Equi, Michael and Finn, Chelsea and Fusai, Niccolo and Groom, Lachy and Hausman, Karol and Ichter, Brian and others},
  journal={arXiv preprint arXiv:2410.24164},
  year={2024}
}

@article{physicalintelligence2025pi05,
  title={$\pi_{0.5}$: A vision-language-action model with open-world generalization},
  author={Black, Kevin and Brown, Noah and Darpinian, James and Dhabalia, Karan and Driess, Danny and Esmail, Adnan and Equi, Michael and Finn, Chelsea and Fusai, Niccolo and Galliker, Manuel Y. and others},
  journal={arXiv preprint arXiv:2504.16054},
  year={2025}
}

@article{smolvla2025,
  title={{SmolVLA}: A vision-language-action model for affordable and efficient robotics},
  author={Shukor, Mustafa and Aubakirova, Dana and Capuano, Francesco and Kooijmans, Pepijn and Palma, Steven and Zouitine, Adil and Aractingi, Michel and Pascal, Caroline and Russi, Martino and Marafioti, Andres and others},
  journal={arXiv preprint arXiv:2506.01844},
  year={2025}
}

@article{scholz2025warm,
  title={Warm starts accelerate conditional diffusion},
  author={Scholz, Jonas and Turner, Richard E.},
  journal={arXiv preprint arXiv:2507.09212},
  year={2025}
}

@article{wagenmaker2025dsrl,
  title={Steering your diffusion policy with latent space reinforcement learning},
  author={Wagenmaker, Andrew and Nakamoto, Mitsuhiko and Zhang, Yunchu and Park, Seohong and Yagoub, Waleed and Nagabandi, Anusha and Gupta, Abhishek and Levine, Sergey},
  journal={arXiv preprint arXiv:2506.15799},
  year={2025}
}

@article{chen2025pirl,
  title={$\pi_{\mathrm{RL}}$: Online {RL} fine-tuning for flow-based vision-language-action models},
  author={Chen, Kang and Liu, Zhihao and Zhang, Tonghe and Guo, Zhen and Xu, Si and Lin, Hao and Zang, Hongzhi and Zhang, Quanlu and Yu, Zhaofei and Fan, Guoliang and Huang, Tiejun and Wang, Yu and Yu, Chao},
  journal={arXiv preprint arXiv:2510.25889},
  year={2025}
}

@inproceedings{florence2022implicit,
  title={Implicit behavioral cloning},
  author={Florence, Pete and Lynch, Corey and Zeng, Andy and Ramirez, Oscar A. and Wahid, Ayzaan and Downs, Laura and Wong, Adrian and Lee, Johnny and Mordatch, Igor and Tompson, Jonathan},
  booktitle={Proceedings of the Conference on Robot Learning},
  year={2022}
}

@article{singh2024revisiting,
  title={Revisiting energy based models as policies: Ranking noise contrastive estimation and interpolating energy models},
  author={Singh, Sumeet and Tu, Stephen and Sindhwani, Vikas},
  journal={Transactions on Machine Learning Research},
  year={2024}
}

@article{inagaki2016coherent,
  title={A coherent {Ising} machine for 2000-node optimization problems},
  author={Inagaki, Takahiro and Haribara, Yoshitaka and Igarashi, Koji and Sonobe, Tomohiro and Tamate, Shuhei and Honjo, Toshimori and Marandi, Alireza and McMahon, Peter L. and Umeki, Takeshi and Enbutsu, Koji and Tadanaga, Osamu and Takenouchi, Hirokazu and Aihara, Kazuyuki and Kawarabayashi, Ken-Ichi and Inoue, Kyo and Utsunomiya, Shoko and Takesue, Hiroki},
  journal={Science},
  volume={354},
  number={6312},
  pages={603--606},
  year={2016},
  publisher={American Association for the Advancement of Science}
}

@article{bohm2022noise,
  title={Noise-injected analog {Ising} machines enable ultrafast statistical sampling and machine learning},
  author={B{\"o}hm, Fabian and Alonso-Urquijo, Diego and Verschaffelt, Guy and Van der Sande, Guy},
  journal={Nature Communications},
  volume={13},
  pages={5847},
  year={2022},
  publisher={Nature Publishing Group}
}

@article{amin2018quantum,
  title={Quantum {Boltzmann} machine},
  author={Amin, Mohammad H. and Andriyash, Evgeny and Rolfe, Jason and Kulchytskyy, Bohdan and Melko, Roger},
  journal={Physical Review X},
  volume={8},
  number={2},
  pages={021050},
  year={2018},
  publisher={American Physical Society}
}

@article{benedetti2017quantum,
  title={Quantum-assisted learning of hardware-embedded probabilistic graphical models},
  author={Benedetti, Marcello and Realpe-G{\'o}mez, John and Biswas, Rupak and Perdomo-Ortiz, Alejandro},
  journal={Physical Review X},
  volume={7},
  number={4},
  pages={041052},
  year={2017},
  publisher={American Physical Society}
}

@article{khoshaman2018quantum,
  title={Quantum variational autoencoder},
  author={Khoshaman, Amir and Vinci, Walter and Denis, Brandon and Andriyash, Evgeny and Sadeghi, Hossein and Amin, Mohammad H.},
  journal={Quantum Science and Technology},
  volume={4},
  number={1},
  pages={014001},
  year={2018},
  publisher={IOP Publishing}
}

@article{vinci2020path,
  title={A path towards quantum advantage in training deep generative models with quantum annealers},
  author={Vinci, Walter and Buffoni, Lorenzo and Sadeghi, Hossein and Khoshaman, Amir and Andriyash, Evgeny and Amin, Mohammad H.},
  journal={Machine Learning: Science and Technology},
  volume={1},
  number={4},
  pages={045028},
  year={2020},
  publisher={IOP Publishing}
}

@article{wilson2021quantum,
  title={Quantum-assisted associative adversarial network: Applying quantum annealing in deep learning},
  author={Wilson, Max and Vandal, Thomas and Hogg, Tad and Rieffel, Eleanor G.},
  journal={Quantum Machine Intelligence},
  volume={3},
  pages={19},
  year={2021},
  publisher={Springer}
}

@article{wang2025quantum,
  title={Quantum-boosted high-fidelity deep learning},
  author={Wang, Feng-ao and Chen, Shaobo and Xuan, Yao and Liu, Junwei and Gao, Qi and Zhu, Hongdong and Hou, Junjie and Yuan, Lixin and Cheng, Jinyu and Yi, Chenxin and Wei, Hai and Ma, Yin and Xu, Tao and Wen, Kai and Li, Yixue},
  journal={arXiv preprint arXiv:2508.11190},
  year={2025}
}

@inproceedings{jia2026a2a,
  title={Action-to-action flow matching},
  author={Jia, Jindou and Li, Gen and Chen, Xiangyu and An, Tuo and Hu, Yuxuan and Li, Jingliang and Guo, Xinying and Yang, Jianfei},
  booktitle={Proceedings of Robotics: Science and Systems},
  year={2026}
}

@inproceedings{Fei_2026_CVPR,
  title={{LIBERO-Plus}: A progressive robustness benchmark for vision-language-action models},
  author={Fei, Senyu and Wang, Siyin and Shi, Junhao and Dai, Zihao and Cai, Jikun and Qian, Pengfang and Ji, Li and He, Xinzhe and Zhang, Shiduo and Fei, Zhaoye and Fu, Jinlan and Gong, Jingjing and Qiu, Xipeng},
  booktitle={Proceedings of the IEEE/CVF Conference on Computer Vision and Pattern Recognition},
  year={2026}
}

@article{wei2026cim,
  title={A versatile coherent {Ising} computing platform},
  author={Wei, Hai and Ai, Chengjun and Guo, Putuo and Jia, Bingjie and Yuan, Lixin and Song, Hanquan and Chen, Shaobo and Cao, Chongyu and Wu, Jie and Ju, Chao and Ma, Yin and Fan, Jintao and Hu, Minglie and Wang, Chuan and Wen, Kai},
  journal={Light: Science \& Applications},
  volume={15},
  number={1},
  pages={74},
  year={2026},
  publisher={Nature Publishing Group}
}

@article{zhai2025vfp,
  title={{VFP}: Variational flow-matching policy for multi-modal robot manipulation},
  author={Zhai, Xuanran and Zhao, Qianyou and Yu, Qiaojun and Hao, Ce},
  journal={arXiv preprint arXiv:2508.01622},
  year={2025}
}

@article{du2026cfvla,
  title={{CF-VLA}: Efficient coarse-to-fine action generation for vision-language-action policies},
  author={Du, Fan and Yan, Feng and Wu, Jianxiong and Xu, Xinrun and Zhang, Weiye and Wang, Weinong and Guo, Yu and Qian, Bin and He, Zhihai and Wang, Fei and Yang, Heng},
  journal={arXiv preprint arXiv:2604.24622},
  year={2026}
}

@article{jia2026mars,
  title={{MARS} policy: Multimodality only when it matters},
  author={Jia, Jindou and An, Tuo and Hu, Yuxuan and Li, Gen and Li, Jingliang and Hou, Bohan and Chen, Xiangyu and Bai, Jiaqi and Lyu, Bofan and Yang, Jianfei},
  journal={arXiv preprint arXiv:2605.29766},
  year={2026}
}

@article{dong2025vitavla,
  title={{VITA-VLA}: Efficiently teaching vision-language models to act via action expert distillation},
  author={Dong, Shaoqi and Fu, Chaoyou and Gao, Haihan and Zhang, Yi-Fan and Yan, Chi and Wu, Chu and Liu, Xiaoyu and Shen, Yunhang and Huo, Jing and Jiang, Deqiang and others},
  journal={arXiv preprint arXiv:2510.09607},
  year={2025}
}

@article{liang2025discretediffusion,
  title={Discrete diffusion {VLA}: Bringing discrete diffusion to action decoding in vision-language-action policies},
  author={Liang, Zhixuan and Li, Yizhuo and Yang, Tianshuo and Wu, Chengyue and Mao, Sitong and Pei, Liuao and Nian, Tian and Zhou, Shunbo and Yang, Xiaokang and Pang, Jiangmiao and Mu, Yao and Luo, Ping},
  journal={arXiv preprint arXiv:2508.20072},
  year={2025}
}

@article{starvla2026,
  title={{StarVLA}: A {Lego}-like codebase for vision-language-action model developing},
  author={{StarVLA Community}},
  journal={arXiv preprint arXiv:2604.05014},
  year={2026}
}

@article{qwen2026qwenvla,
  title={{Qwen-VLA}: Unifying vision-language-action modeling across tasks, environments, and robot embodiments},
  author={Wang, Qiuyue and Li, Mingsheng and Guan, Jian and Ye, Jinhui and Xie, Sicheng and Liu, Yitao and Chen, Junhao and Liang, Zhixuan and Zhang, Jie and Hu, Xintong and others},
  journal={arXiv preprint arXiv:2605.30280},
  year={2026}
}

@article{zheng2026pokevla,
  title={{PokeVLA}: Empowering pocket-sized vision-language-action model with comprehensive world knowledge guidance},
  author={Zheng, Yupeng and Li, Xiang and Gu, Songen and Zheng, Yuhang and Tian, Shuai and Li, Weize and Wang, Linbo and Fei, Senyu and Li, Pengfei and Gao, Yinfeng and others},
  journal={arXiv preprint arXiv:2604.20834},
  year={2026}
}

@article{sun2026vlajepa,
  title={{VLA-JEPA}: Enhancing vision-language-action model with latent world model},
  author={Sun, Jingwen and Zhang, Wenyao and Qi, Zekun and Ren, Shaojie and Liu, Zezhi and Zhu, Hanxin and Sun, Guangzhong and Jin, Xin and Chen, Zhibo},
  journal={arXiv preprint arXiv:2602.10098},
  year={2026}
}

@inproceedings{zhong2026acotvla,
  title={{ACoT-VLA}: Action chain-of-thought for vision-language-action models},
  author={Zhong, Linqing and Liu, Yi and Wei, Yifei and Xiong, Ziyu and Yao, Maoqing and Liu, Si and Ren, Guanghui},
  booktitle={Proceedings of the IEEE/CVF Conference on Computer Vision and Pattern Recognition},
  year={2026}
}

@article{jia2026ode,
  title={{ODE} policy: Robot learning via action dynamics, not actions},
  author={Jia, Jindou and Li, Jingliang and Zuo, Kuangji and An, Tuo and Hu, Yuxuan and Zhao, Mengfei and Lyu, Bofan and Bai, Jiaqi and Zhu, Daoyuan and Dai, Hengyu and Zhang, Junpeng and Yang, Jianfei},
  journal={arXiv preprint arXiv:2611.01729},
  year={2026}
}

@article{chen2025robotwin20,
  title={{RoboTwin 2.0}: A Scalable Data Generator and Benchmark with Strong Domain Randomization for Robust Bimanual Robotic Manipulation},
  author={Chen, Tianxing and Chen, Zanxin and Chen, Baijun and Cai, Zijian and Liu, Yibin and Li, Zixuan and Liang, Qiwei and Lin, Xianliang and Ge, Yiheng and Gu, Zhenyu and Deng, Weiliang and Guo, Yubin and Nian, Tian and Xie, Xuanbing and Chen, Qiangyu and Su, Kailun and Xu, Tianling and Liu, Guodong and Hu, Mengkang and Gao, Huan-ang and Wang, Kaixuan and Liang, Zhixuan and Qin, Yusen and Yang, Xiaokang and Luo, Ping and Mu, Yao},
  journal={arXiv preprint arXiv:2506.18088},
  year={2025},
  doi={10.48550/arXiv.2506.18088}
}

@misc{robotwin2026leaderboard,
  title={{RoboTwin 2.0} Leaderboard},
  author={{RoboTwin Team}},
  howpublished={\url{https://robotwin-platform.github.io/leaderboard}},
  year={2026},
  note={Accessed November 17, 2026}
}

@article{yang2026eventvla,
  title={{EventVLA}: Event-Driven Visual Evidence Memory for Long-Horizon Vision-Language-Action Policies},
  author={Yang, Ganlin and Tu, Zhangzheng and Yang, Yuqiang and Mao, Sitong and Dong, Junyi and Chen, Tianxing and Peng, Jiaqi and Xiong, Jing and Cao, Jiafei and Dai, Jifeng and Zhou, Wengang and Mu, Yao and Wang, Tai},
  journal={arXiv preprint arXiv:2606.20092},
  year={2026},
  doi={10.48550/arXiv.2606.20092}
}

@inproceedings{shou2026halo,
  title={{HALO}: A unified vision-language-action model for embodied multimodal chain-of-thought reasoning},
  author={Shou, Quanxin and Zhu, Fangqi and Chen, Shuang and Yan, Puxin and Yan, Zhengyang and Miao, Yikun and Pang, Xiaoyi and Hong, Zicong and Shi, Ruikai and Huang, Hao and Zhang, Jie and Guo, Song},
  booktitle={Proceedings of the International Conference on Machine Learning},
  year={2026}
}

@inproceedings{li2026spatialforcing,
  title={Spatial forcing: Implicit spatial representation alignment for vision-language-action model},
  author={Li, Fuhao and Song, Wenxuan and Zhao, Han and Wang, Jingbo and Ding, Pengxiang and Wang, Donglin and Zeng, Long and Li, Haoang},
  booktitle={Proceedings of the International Conference on Learning Representations},
  year={2026}
}

@article{dai2026leap,
  title={Where should action generation begin? A learnable source prior for generative robot policies},
  author={Dai, Meipo and Zhuang, Qiyuan and Xu, He-Yang and Shuai, Ying-Jie and Wang, Yijun and Dou, Qi and Wei, Xiu-Shen},
  journal={arXiv preprint arXiv:2606.17408},
  year={2026}
}

@inproceedings{gao2026vita,
  title={{VITA}: Vision-to-action flow matching policy},
  author={Gao, Dechen and Zhao, Boqi and Lee, Andrew and Chuang, Ian and Zhou, Hanchu and Wang, Hang and Zhao, Zhe and Zhang, Junshan and Soltani, Iman},
  booktitle={Proceedings of the International Conference on Learning Representations},
  year={2026}
}

@inproceedings{sochopoulos2025cot,
  title={Fast flow-based visuomotor policies via conditional optimal transport couplings},
  author={Sochopoulos, Andreas and Malkin, Nikolay and Tsagkas, Nikolaos and Moura, Jo{\~a}o and Gienger, Michael and Vijayakumar, Sethu},
  booktitle={Proceedings of the Conference on Robot Learning},
  year={2025}
}

@inproceedings{feng2026gridsampler,
  title={See what matters: Differentiable grid sample pruning for generalizable vision-language-action model},
  author={Feng, Yixu and Zhao, Zinan and Ma, Yanxiang and Xia, Chenghao and Du, Chengbin and Wang, Yunke and Xu, Chang},
  booktitle={Proceedings of the International Conference on Machine Learning},
  year={2026}
}

@article{shi2026memoryvlapp,
  title={{MemoryVLA++}: Temporal modeling via memory and imagination in vision-language-action models},
  author={Shi, Hao and Li, Weiye and Xie, Bin and Wang, Yulin and Zhou, Renping and Wang, Tiancai and Zhang, Xiangyu and Luo, Ping and Huang, Gao},
  journal={arXiv preprint arXiv:2606.09827},
  year={2026}
}

@inproceedings{yuan2026fastwam,
  title={{Fast-WAM}: Do World Action Models Need Test-time Future Imagination?},
  author={Yuan, Tianyuan and Dong, Zibin and Liu, Yicheng and Zhao, Hang},
  booktitle={Advances in Neural Information Processing Systems},
  year={2026},
  note={Accepted}
}

@inproceedings{lin2026evo1,
  title={{Evo-1}: Lightweight Vision-Language-Action Model with Preserved Semantic Alignment},
  author={Lin, Tao and Zhong, Yilei and Du, Yuxin and Zhang, Jingjing
          and Liu, Jiting and Chen, Yinxinyu and Gu, Encheng and Liu, Ziyan
          and Cai, Hongyi and Zou, Yanwen and Zou, Lixing and Zhou, Zhaoye
          and Li, Gen and Zhao, Bo},
  booktitle={Proceedings of the IEEE/CVF Conference on Computer Vision and Pattern Recognition},
  pages={13397--13406},
  year={2026}
}

@inproceedings{huang2026notvla,
  title={{NoTVLA}: Semantics-Preserving Robot Adaptation via Narrative Action Interfaces},
  author={Huang, Zheng and Liu, Mingyu and Lin, Xiaoyi and Zhu, Muzhi
          and Zhao, Canyu and Du, Zongze and Lin, Ye and Li, Xiaoman
          and Jia, Yiduo and Zhong, Hao and Chen, Hao and Shen, Chunhua},
  booktitle={Advances in Neural Information Processing Systems},
  year={2026},
  note={Accepted}
}
}

\end{document}